\documentclass[conference]{IEEEtran}
\newif\ifarxiv
\arxivtrue

\usepackage{cite}
\usepackage{amsmath,amssymb,amsfonts}
\usepackage{algorithmic}
\usepackage{graphicx}
\usepackage{textcomp}
\usepackage{xcolor}
\def\BibTeX{{\rm B\kern-.05em{\sc i\kern-.025em b}\kern-.08em
    T\kern-.1667em\lower.7ex\hbox{E}\kern-.125emX}}

\usepackage{relsize}

\usepackage{booktabs}
\usepackage{moreverb}
\usepackage{mathtools}
\usepackage[final]{pdfpages}
\usepackage{float}
\usepackage{scalefnt}
\usepackage{enumitem}
\usepackage[linesnumbered,ruled,vlined]{algorithm2e}
\usepackage{colortbl}
\usepackage{xfrac}
\usepackage{multicol, multirow, hhline}
\usepackage{diagbox}
\usepackage{adjustbox}

\usepackage{subcaption}
\ifarxiv
\definecolor{iccvblue}{rgb}{0.21,0.49,0.74}
\usepackage[hidelinks, breaklinks,colorlinks,allcolors=iccvblue]{hyperref}
\else
\usepackage[hidelinks]{hyperref}       % hyperlinks
\fi

\usepackage[capitalize]{cleveref}
\crefname{equation}{}{}

\usepackage{afterpage}
\usepackage{tikz}
\usetikzlibrary{matrix, arrows, backgrounds,shapes.geometric,shapes.arrows,decorations.pathreplacing}

\definecolor{CarnationPink}{RGB}{255,166,201}
\definecolor{Yellow}{RGB}{255,255,0}
\definecolor{PastelGreen}{RGB}{177,221,159}
\definecolor{Green1}{RGB}{138,255,103}
\definecolor{Green2}{RGB}{50,190,31}
\definecolor{Green3}{RGB}{0,165,138}
\definecolor{PastelPurple}{RGB}{223,192,243}
\definecolor{PastelOrange}{RGB}{250,207,157}
\definecolor{PastelBlue}{RGB}{184,219,252}
\definecolor{LightGray}{gray}{0.9}

\ifarxiv
\usepackage[absolute,overlay]{textpos}
\fi
    
\begin{document}

\title{Deep Learning for Unrelated-Machines Scheduling: Handling Variable Dimensions}

\iffalse
\author{Anonymous Authors\\
  Affiliations Withheld for Anonymous Review}
\fi
%\iffalse
\author{\IEEEauthorblockN{Diego {Hitzges}}
\IEEEauthorblockA{\textit{Dept. of Mathematics} \\
\textit{Technische Universität Berlin} \\
Berlin, Germany \\
\href{https://orcid.org/0009-0000-3978-0591}{0009-0000-3978-0591}
}
\and
\IEEEauthorblockN{Guillaume Sagnol}
\IEEEauthorblockA{\textit{Dept. of Mathematics} \\
\textit{Technische Universität Berlin}\\
Berlin, Germany \\
\href{https://orcid.org/0000-0001-6910-8907}{0000-0001-6910-8907}
}
}
%\fi

\maketitle

\ifarxiv
% Command to write a header to say "paper accepted at such conference"
%\usepackage{color}
\definecolor{somegray}{gray}{0.5}
\newcommand{\darkgrayed}[1]{\textcolor{somegray}{#1}}
\begin{textblock}{11}(2.5, 0.2)
\begin{center}
\darkgrayed{This paper has been accepted for publication at the\\
24th IEEE International Conference on Machine Learning and Applications (ICMLA 2025).}
\end{center}
\end{textblock}
\fi

\begin{abstract}
Deep learning has been effectively applied to many discrete optimization problems. However, learning-based scheduling on unrelated parallel machines remains particularly difficult to design. Not only do the numbers of jobs and machines vary, but each job-machine pair has a unique processing time, dynamically altering feature dimensions. We propose a novel approach with a neural network tailored for offline deterministic scheduling of arbitrary sizes on unrelated machines. The goal is to minimize a complex objective function that includes the makespan and the weighted tardiness of jobs and machines. Unlike existing online approaches, which process jobs sequentially, our method generates a complete schedule considering the entire input at once. The key contribution of this work lies in the sophisticated architecture of our model. By leveraging various NLP-inspired architectures, it effectively processes any number of jobs and machines with varying feature dimensions imposed by unrelated processing times. Our approach enables supervised training on small problem instances while demonstrating strong generalization to much larger scheduling environments. Trained and tested on instances with 8 jobs and 4 machines, costs were only 2.51\% above optimal. Across all tested configurations of up to 100 jobs and 10 machines, our network consistently outperformed an advanced dispatching rule, which incurred 22.22\% higher costs on average. As our method allows fast retraining with simulated data and adaptation to various scheduling conditions, we believe it has the potential to become a standard approach for learning-based scheduling on unrelated machines and similar problem environments.
\end{abstract}
\section{Introduction}

Discrete optimization problems are not only of mathematical interest but also of practical relevance due to their impact on real-world applications. One particularly promising category for increasing efficiency is job scheduling. In this context, tasks (\emph{jobs})
are assigned to suitable resources (\emph{machines}) with the aim to minimize the costs dictated by a specific objective function. Like many combinatorial problems, job scheduling quickly becomes NP-hard as complexity increases~\cite{pinedo2008scheduling}. Approximation algorithms have mostly been developed for simplified scenarios and 
%are typically 
tailored to narrowly defined problem settings, thus lacking generalizability. Deep learning presents a promising avenue for addressing these challenges. However, traditional neural network structures often struggle with the variability in input and output dimensions inherent to scheduling problems. Such flexibility is particularly crucial given the fluctuation in numbers of jobs and machines, especially in offline settings where all data must be processed simultaneously. One field in which the need for such adaptability has been effectively addressed is Natural Language Processing (NLP). 
%Therefore, 
In this paper, we introduce a novel network architecture that integrates various NLP-inspired structures and provides the flexibility of processing any number of jobs and machines, while also handling varying feature dimensions. By applying this architecture, we provide an offline solution to a complex scheduling environment that can also 
%potentially 
be adapted to different problem settings. In doing so, we aim to overcome the shortcomings in both mathematical and learning based approaches.

\subsection{Problem Formulation}

We tackle the deterministic offline problem of nonpreemptively scheduling $J$ jobs on $M$ unrelated machines, denoted by $R_m$. Each job $j$ has a specific processing time $p_{jm}>0$ on each machine $m$, along with an individual deadline $d_j$ and penalty weight $w_j$ for exceeding this deadline. Similarly, each machine $m$ has a deactivation deadline $\delta_m$ and an associated penalty weight $\omega_m$. Additionally, machines begin with initial occupations of runtime $r_m$. The objective is to minimize the combined total of the makespan $C_{max}$ with the total weighted tardiness of the jobs $T_j$ and machines $\tau_m$. This leads us to the problem formulation:
\begin{equation}
\resizebox{0.9\columnwidth}{!}{$
R_m \mid d_j,\delta_m, w_j, \omega_m, r_m \mid C_{\max} + \sum_{j=1}^J w_j T_j + \sum_{m=1}^M \omega_m \tau_m .
$}
\label{our_JSP}
\end{equation}

\subsection{Our Approach}

To effectively apply deep learning to our complex scheduling problem, we embed it within a framework conducive to neural network operations. We achieve this by reformulating the task as a deterministic Markov Decision Problem (MDP). In this framework, each moment a machine becomes available is represented as a state, necessitating an action to either assign a job to it or deactivate it. Starting from the initial state, which represents the given problem environment, a schedule is systematically generated by iteratively applying the neural network to estimate the optimal action at each state. As each state incorporates comprehensive data about all pending jobs and active machines, this approach functions effectively as an offline solution. Moreover, due to the Markovian properties of our framework, each subsequent state forms its own distinct scheduling scenario, thus allowing for online applications and modifications as well. 
We will demonstrate that our network effectively generalizes to significantly larger-scale problems than those on which it was trained. While our framework is compatible with Reinforcement Learning (RL), our network’s remarkable generalization capabilities allow us to train it effectively using supervised learning on only very small instances, where exact state-action values are inexpensive to compute. This enables both high-precision decision-making and rapid retraining for new problem settings.
%This is a key advantage, as it ensures the model learns from fully accurate optimal policies, enabling both high-precision decision-making and rapid retraining for new problem settings.
The trained model and source code are publicly accessible in an online repository
%~\cite{anonymous_repo}
~\cite{hitzges2025scheduling}
to support clarity, reproducibility and practical applicability.
\section{Related Work}

Our work addresses an unrelated-machines scheduling problem (\ref{our_JSP}) with a compound objective function, focusing on minimizing both the  makespan $C_{max}$, which is already NP-hard on two identical machines, and the total weighted tardiness of jobs on unrelated machines, a problem that poses significant challenges in even modestly sized environments of around 5 machines and 30 jobs~\cite[pages 112, 143]{pinedo2008scheduling}. In addition to exact solution approaches based on Dynamic Programming or Mixed-Integer Programming (which can only handle very small problems), several heuristic approaches have thus been proposed to attack this class of problems; see~\cite{durasevic2023heuristic} and the references therein. While these heuristics are generally based on dispatching rules, which order jobs using a priority function; see, e.g.,~\cite{yang2013dispatching}, or evolutionary algorithms, such as~\cite{zhou2007scheduling} who use an ant colony optimization algorithm, learning-based algorithms have begun to emerge.

The majority of existing literature on learning-based scheduling predominantly employs RL or Deep Reinforcement Learning (DRL), as for example~\cite{bouvska2023deep} who used deep learning to minimize the total tardiness on a single machine. Even though compatible with DRL, our framework adopts  supervised learning due to its quicker convergence and higher accuracy, providing a potentially more versatile solution that can be swiftly retrained to be adapted for varying problem scenarios. Supervised Learning was also utilized in~\cite{skadel2017online}, yet only online in flow-shop problems. Several studies have been done on job-shop or flow-shop problems~\cite{kumar2019multiple,mao2016resource,tassel2021reinforcement}. In fact, a comprehensive literature review to RL for machine scheduling found that 50 out of 80 selected publications were on these topics~\cite{kayhan2023reinforcement}. From the remaining 30 studies, only few focused on problems with unrelated machines. All of them circumvent the problems of having to deal with a state-space and an action-space of varying size, either by feeding information on only one of the remaining jobs to the neural network~\cite{abraham2019reinforcement}, or by aggregating the state information to a handful of features and restricting the action space to the selection of a dispatching rule (rather than the exact identification of which job should be started)~\cite{zhang2012minimizing,guo2020optimization}. Moreover, many review papers and surveys on learning based methods for combinatorial optimization problems tend to cover RL applications in production systems or offer general insights on how DRL can be applied to different network architectures for combinatorial problems, without detailing their application to machine scheduling problems~\cite{khadivi2023deep}. This highlights a noticeable gap of adaptable yet specific learning-based solutions to complex unrelated-machines scheduling problems, which this paper tries to bridge.
\section{Methodology}\label{section:method}

In this section, we outline the transformation of our scheduling problem (\ref{our_JSP}) into an MDP framework suitable for deep learning applications. We detail the training process, describe the metrics, and, in the absence of a benchmark, construct an advanced dispatching rule with a priority function tailored to our specific compound objective function to evaluate our network’s performance. 

\subsection{States as Inputs}\label{section:states_as_inputs}

% Define colors from colortbl
\definecolor{CarnationPink}{RGB}{255,166,201}
\definecolor{Yellow}{RGB}{255,255,0}

% Define colors from colortbl
\definecolor{CarnationPink}{RGB}{255,166,201}
\definecolor{Yellow}{RGB}{255,255,0}

\definecolor{PastelGreen}{RGB}{177,221,159}

\definecolor{Green1}{RGB}{138,255,103}
\definecolor{Green2}{RGB}{50,190,31}
\definecolor{Green3}{RGB}{0,165,138}

\definecolor{PastelPurple}{RGB}{223,192,243}
\definecolor{PastelOrange}{RGB}{250,207,157}
\definecolor{PastelBlue}{RGB}{184,219,252}
\definecolor{LightGray}{gray}{0.9}

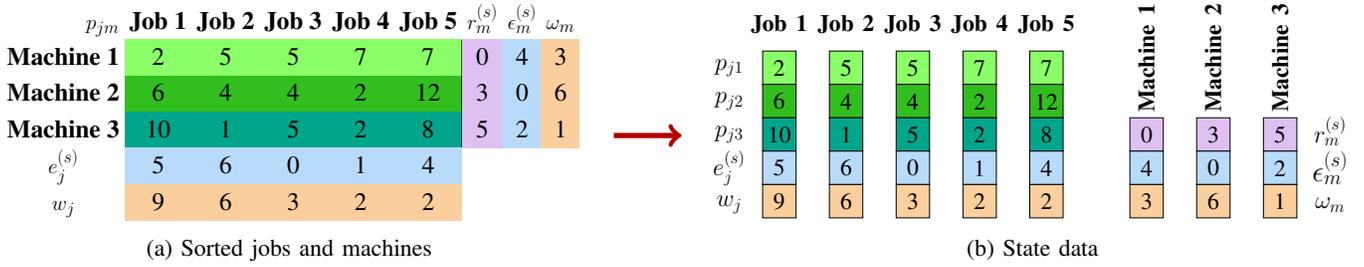
\begin{figure*}
    \centering
    \begin{subfigure}[b]{0.43\textwidth}%{0.43\textwidth}
        \centering
        \resizebox{\textwidth}{!}{%
            {\scalefont{3}%
%\newcolumntype{a}{>{\columncolor{PastelOrange}}c}
\renewcommand{\arraystretch}{1.2}
\begin{tabular}{c c c c c c c c c}
\multicolumn{1}{r}{\textit{$p_{jm}$}}                         & \textbf{Job 1} & \textbf{Job 2} & \textbf{Job 3} & \textbf{Job 4} & \textbf{Job 5}          & \textit{$r_m^{(s)}$} & \textit{$\epsilon_m^{(s)}$} & \textit{$\omega_m$} \\ %\hhline{~*{8}{-}} 

\multicolumn{1}{c}{\textbf{Machine 1 }}   & \cellcolor{Green1}2              & \cellcolor{Green1}5              & \cellcolor{Green1}5              & \cellcolor{Green1}7              & \multicolumn{1}{c|}{\cellcolor{Green1}7}  & \cellcolor{PastelPurple}0                    & \cellcolor{PastelBlue}4                           & \cellcolor{PastelOrange}3                   \\

\multicolumn{1}{c}{\textbf{Machine 2 }}   & \cellcolor{Green2}6              & \cellcolor{Green2}4              & \cellcolor{Green2}4              & \cellcolor{Green2}2              & \multicolumn{1}{c|}{\cellcolor{Green2}12} & \cellcolor{PastelPurple}3                    & \cellcolor{PastelBlue}0                           & \cellcolor{PastelOrange}6 
                  \\
\multicolumn{1}{c}{\textbf{Machine 3 }}   & \cellcolor{Green3}10             & \cellcolor{Green3}1              & \cellcolor{Green3}5              & \cellcolor{Green3}2              & \multicolumn{1}{c|}{\cellcolor{Green3}8}  & \cellcolor{PastelPurple}5                    & \cellcolor{PastelBlue}2                           & \cellcolor{PastelOrange}1                   \\ %\hhline{~*{5}{-}}
 
\multicolumn{1}{c}{\textit{$e_j^{(s)}$}} & \cellcolor{PastelBlue}5              & \cellcolor{PastelBlue}6              & \cellcolor{PastelBlue}0              & \cellcolor{PastelBlue}1              & \cellcolor{PastelBlue}4   &                      &                             &                     \\

\multicolumn{1}{c}{\textit{$w_j$}}       & \cellcolor{PastelOrange}9              & \cellcolor{PastelOrange}6              & \cellcolor{PastelOrange}3              & \cellcolor{PastelOrange}2              & \cellcolor{PastelOrange}2  &                      &                             &                    
\end{tabular}}
        }
        \caption{Sorted jobs and machines}
        \label{fig_a:state to data}
    \end{subfigure}
    \hfill
    \begin{tikzpicture}[overlay, remember picture]
        \draw[->, thick, red!70!black, line width=0.7mm] (-0.4, 1.7) -- (0.5, 1.7);
    \end{tikzpicture}
    \hfill
    %\vspace{10pt}
    \begin{subfigure}[b]{0.48\textwidth}%{0.48\textwidth}
        \centering
        \resizebox{\textwidth}{!}{%
        {\scalefont{2}%
\begin{tikzpicture}

%Machine 1
\node[align=center, rotate=90] at (11.5,-0.2) {\fontseries{sb}\selectfont\huge{Machine 1}};
\draw (11,-2) [fill=PastelPurple]rectangle ++(1,-1) node[midway]{0};
\draw (11,-3) [fill=PastelBlue]rectangle ++(1,-1) node[midway]{4};
\draw (11,-4) [fill=PastelOrange]rectangle ++(1,-1) node[midway]{3};

%Machine 2
\node[align=center, rotate=90] at (13.5,-0.2) {\fontseries{sb}\selectfont\huge{Machine 2}};
\draw (13,-2) [fill=PastelPurple]rectangle ++(1,-1) node[midway]{3};
\draw (13,-3) [fill=PastelBlue]rectangle ++(1,-1) node[midway]{0};
\draw (13,-4) [fill=PastelOrange]rectangle ++(1,-1) node[midway]{6};

%Machine 3
\node[align=center, rotate=90] at (15.5,-0.2) {\fontseries{sb}\selectfont\huge{Machine 3}};
\draw (15,-2) [fill=PastelPurple]rectangle ++(1,-1) node[midway]{5};
\draw (15,-3) [fill=PastelBlue]rectangle ++(1,-1) node[midway]{2};
\draw (15,-4) [fill=PastelOrange]rectangle ++(1,-1) node[midway]{1};

%Name entries to the right of Machine 3
\node at (17,-2.4) {\relsize{+0.9}{$r^{(s)}_m$}};
\node at (17,-3.5) {\relsize{+1.3}{$\epsilon^{(s)}_m$}};
\node at (17,-4.6) {\relsize{+0.9}{$\omega_m$}};

%Name entries to the left of Job 1
\node at (-1,-0.5) {\relsize{+0.9}{$p_{j1}$}};
\node at (-1,-1.5) {\relsize{+0.9}{$p_{j2}$}};
\node at (-1,-2.5) {\relsize{+0.9}{$p_{j3}$}};
\node at (-1,-3.5) {\relsize{+0.9}{$e^{(s)}_j$}};
\node at (-1,-4.6) {\relsize{+0.9}{$w_j$}};

%Job 1
\node[align=center] at (0.5,0.8) {\fontseries{sb}\huge{Job 1}};
\draw (0,0) [fill=Green1]rectangle ++(1,-1) node[midway]{2};
\draw (0,-1) [fill=Green2]rectangle ++(1,-1) node[midway]{6};
\draw (0,-2) [fill=Green3]rectangle ++(1,-1) node[midway]{10};
\draw (0,-3) [fill=PastelBlue]rectangle ++(1,-1) node[midway]{5};
\draw (0,-4) [fill=PastelOrange]rectangle ++(1,-1) node[midway]{9};

%Job 2
\node[align=center] at (2.5,0.8) {\fontseries{sb}\huge{Job 2}};
\draw (2,0) [fill=Green1]rectangle ++(1,-1) node[midway]{5};
\draw (2,-1) [fill=Green2]rectangle ++(1,-1) node[midway]{4};
\draw (2,-2) [fill=Green3]rectangle ++(1,-1) node[midway]{1};
\draw (2,-3) [fill=PastelBlue]rectangle ++(1,-1) node[midway]{6};
\draw (2,-4) [fill=PastelOrange]rectangle ++(1,-1) node[midway]{6};

%Job 3
\node[align=center] at (4.5,0.8) {\fontseries{sb}\huge{Job 3}};
\draw (4,0) [fill=Green1]rectangle ++(1,-1) node[midway]{5};
\draw (4,-1) [fill=Green2]rectangle ++(1,-1) node[midway]{4};
\draw (4,-2) [fill=Green3]rectangle ++(1,-1) node[midway]{5};
\draw (4,-3) [fill=PastelBlue]rectangle ++(1,-1) node[midway]{0};
\draw (4,-4) [fill=PastelOrange]rectangle ++(1,-1) node[midway]{3};

%Job 4
\node[align=center] at (6.5,0.8) {\fontseries{sb}\huge{Job 4}};
\draw (6,0) [fill=Green1]rectangle ++(1,-1) node[midway]{7};
\draw (6,-1) [fill=Green2]rectangle ++(1,-1) node[midway]{2};
\draw (6,-2) [fill=Green3]rectangle ++(1,-1) node[midway]{2};
\draw (6,-3) [fill=PastelBlue]rectangle ++(1,-1) node[midway]{1};
\draw (6,-4) [fill=PastelOrange]rectangle ++(1,-1) node[midway]{2};

%Job 5
\node[align=center] at (8.5,0.8) {\fontseries{sb}\huge{Job 5}};
\draw (8,0) [fill=Green1]rectangle ++(1,-1) node[midway]{7};
\draw (8,-1) [fill=Green2]rectangle ++(1,-1) node[midway]{12};
\draw (8,-2) [fill=Green3]rectangle ++(1,-1) node[midway]{8};
\draw (8,-3) [fill=PastelBlue]rectangle ++(1,-1) node[midway]{4};
\draw (8,-4) [fill=PastelOrange]rectangle ++(1,-1) node[midway]{2};
\end{tikzpicture}}
        }
        \caption{State data}
        \label{fig_b:state to data}
    \end{subfigure}
    %\vspace{7pt}
    \caption{State information to data}
    \label{fig:state to data}
\end{figure*}

States are constructed as inputs for our neural network whenever a machine becomes free during the scheduling process. In each state $s \in S$, machines are alphabetically sorted primarily by remaining runtime, secondarily by higher weight, and tertiarily by earlier deadline. Analogously, a permutation
\begin{equation}
    \Pi_m: \{1, \ldots, J\} \mapsto \{1, \ldots J\} \label{perm:jobs}
\end{equation}
is then applied to the jobs, alphabetically sorting them primarily by processing time on the free machine $m$ with the lowest index, secondarily by higher weight, and tertiarily by earlier deadline. Every state is represented by vectors for each pending job and active machine. Vectors for the machines are three-dimensional, consisting of the remaining runtime $r_m^{(s)}$, \emph{earliness} $\epsilon_m^{(s)}$ --- the time until a machine's deadline --- and weight $\omega_m$. Job vectors include processing times $p_{jm}$ for each active machine $m$, along with the job´s earliness $e_j^{(s)}$ and weight $w_j$. An exemplary illustration of how to extract the data from a state is given by \cref{fig:state to data}.

\subsection{Action Values as Targets}

In our MDP, each state $s \in S$ corresponds to a decision point where a machine becomes available. The set of feasible actions $A(s)$ at each state includes assigning any of the remaining jobs to the machine, or deactivating it as long as another active machine exists, as idle time would always be suboptimal to an immediate assignment. This makes the total number of actions per state equal to the number of pending jobs plus one, if deactivation is feasible. A final state is reached as soon as no pending jobs remain. An intuitive target for our neural network would be the Q-values from the Bellman equation, from which an optimal policy $\mu^*$ can immediately be deduced. However, to shift focus on distinguishing between the best options and simply discarding the bad ones, we define our target values \(y^{(s)}(a)\) by inversely scaling the Q-values with the optimal costs $V^*(s):=V_{\mu^*}(s)$ and applying softmax:
\begin{equation}
y^{(s)}(a) := \text{softmax}(\frac{V^*(s)}{Q^*(s,a)}) \in [0,1].
\label{eq:target_values}
\end{equation}

\subsection{Neural Network as Policy}

The scheduling algorithm generated by our neural network with parameters $\vartheta$ can be applied to any scheduling problem as defined by equation (\ref{our_JSP}). To generate a schedule, transform the problem into an MDP with corresponding initial state $s \in S$ and apply the network-induced policy $\mu_\vartheta$. This involves:
\begin{enumerate}
    \item Feeding the state data into the neural network.
    \item Selecting the action with the highest estimated probability from the output distribution over feasible actions.
    \item Transitioning to the successor state by executing the chosen action.
\end{enumerate}
Repeat this process until a final state is reached. The resulting sequence of action-state pairs forms a feasible schedule, with objective costs equivalent to the policy costs $V_{\mu_\vartheta}$. Since $\mu_\vartheta$ approximates the optimal policy $\mu^*$, the derived schedule provides an estimation of the optimal schedule.

\subsection{Data Simulation and Supervised Training}\label{section:data}

Supervised training is used on our neural network, offering short training times and enhancing the predictive accuracy and reliability of our model. We simulate 100.000 scheduling problems of the form (\ref{our_JSP}) with 8 jobs and 4 machines according to the following parameters:
\begin{itemize}
    \item Maximum weight $w_{\max} = 10$
    \item Maximum deadline: $d_{\max} = 30$ (resembling a month)
    \item Maximum processing time: $p_{\max}$ randomly selected from $[ \frac{1}{3}d_{\max}, \frac{3}{2}d_{\max} ]$
    \item Maximum initial runtime for machines: randomly selected from $[ \frac{1}{3}p_{\max}, p_{\max} ]$
\end{itemize}
Jobs and machines are generated with processing times, deadlines, weights, and runtimes randomly chosen within these ranges. If no machine is initially idle, one is randomly set to zero runtime. For each scheduling problem, we compute all possible states together with their ground-truth action values. We then select 18 of these states, one for each combination of 3 to 8 pending jobs and 2 to 4 active machines. To ensure a balanced dataset, the states are uniformly chosen with respect to their optimal action across feasible actions. The performance of our model with parameters $\vartheta$ is evaluated against the optimal costs by measuring the relative cost increase per state-action pair with the following metric:
\begin{equation}
\Delta Q(s,\mu_\vartheta(s)) := \frac{Q^*(s,\mu_\vartheta(s)) - V^*(s)}{V^*(s)}.
\label{eq:training_metric}
\end{equation}
We opt for the Mean Squared Error (MSE) as loss function over the Cross-Entropy to precisely penalize deviations from near-optimal actions. Training is performed using Python3 with TensorFlow. We use the Adam optimizer with a learning rate of 0.001 and a batch size of 128. The network is trained for 30 epochs on a private laptop equipped with an AMD Ryzen 5 3500 CPU (2.1 GHz 4 cores, 8 threads), 8 GB of RAM and an 80 GB SSD, taking approximately 24 hours.

\subsection{Learning Based versus Heuristic Scheduling} \label{section:heuristic} 

We have developed an advanced dispatching rule with a priority function tailored to our specific compound objective function to evaluate the performance of our neural network on scheduling problems with a variety of numbers of jobs and machines. Dispatching rules, such as those described in~\cite{durasevic2023heuristic}, have been shown to provide effective heuristics for complex scheduling problems. Inspired by greedy list scheduling, which has demonstrated reasonable approximations for several objective functions in scheduling on identical machines~\cite[page 40]{williamson2010design}, we enhance this approach to address the higher complexity of our problem. Whenever a machine $m$ becomes available, jobs are sorted according to $\Pi_m$ (\ref{perm:jobs}). We check whether the first pending job $j_1$ in that sorted list could finish sooner on another active machine $m^{*}$, i.e. if $\exists \ m^* : (r_{m^*} + p_{j_1m^*} < p_{j_1m}).$ If no such machine $m^{*}$ exists, job $j_1$ is assigned to machine $m$. Otherwise, the process is repeated for the next remaining job until a job is assigned. If the iteration completes without an assignment, the free machine $m$ is deactivated.
\section{Network Architecture} \label{section:network}

\tikzset{embed_vec/.style = {matrix of math nodes,left delimiter=(,right delimiter=)},
shorten <>/.style={ shorten >=#1, shorten <=#1 }}

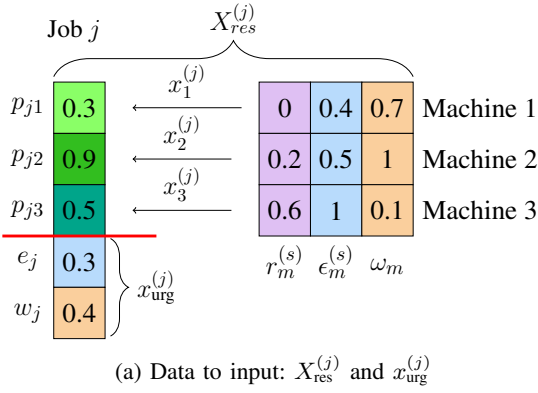
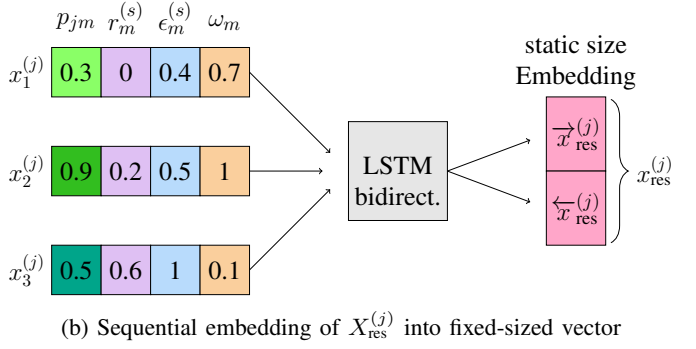
\begin{figure*}[t]
    \centering
    \begin{subfigure}[b]{0.4\textwidth}%{0.3\textwidth}
        \centering
        \resizebox{\textwidth}{!}{%
        {\scalefont{1.5}%
\begin{tikzpicture}

%Job
\node[align=center] at (0.4,1) {\Large{Job $j$}};
\draw (0,0) [fill=Green1]rectangle ++(1,-1) node[midway](P1){0.3};
\draw (0,-1) [fill=Green2]rectangle ++(1,-1) node[midway](P2){0.9};
\draw (0,-2) [fill=Green3]rectangle ++(1,-1) node[midway](P3){0.5};
\draw (0,-3) [fill=PastelBlue]rectangle ++(1,-1) node[midway]{0.3};
\draw (0,-4) [fill=PastelOrange]rectangle ++(1,-1) node[midway]{0.4};

%Job Descriptions
\node[align=center] at (-0.5,-0.5) {$p_{j1}$};
\node[align=center] at (-0.5,-1.5) {$p_{j2}$};
\node[align=center] at (-0.5,-2.5) {$p_{j3}$};
\node[align=center] at (-0.5,-3.5) {$e_j$};
\node[align=center] at (-0.5,-4.5) {$w_j$};

%Machine 1
%\node[align=center] at (3,-0.5) {\Large{Machine 1}};
\draw (4,0) [fill=PastelPurple]rectangle ++(1,-1) node[midway](M1){0};
\draw (5,0) [fill=PastelBlue]rectangle ++(1,-1) node[midway]{0.4};
\draw (6,0) [fill=PastelOrange]rectangle ++(1,-1) node[midway]{0.7};
\node[anchor=west] at (7,-0.5) {Machine 1};

%Machine 2
%\node[align=center] at (3,-1.5) {\Large{Machine 2}};
\draw (4,-1) [fill=PastelPurple]rectangle ++(1,-1) node[midway](M2){0.2};
\draw (5,-1) [fill=PastelBlue]rectangle ++(1,-1) node[midway]{0.5};
\draw (6,-1) [fill=PastelOrange]rectangle ++(1,-1) node[midway]{1};
\node[anchor=west] at (7,-1.5) {Machine 2};

%Machine 3
%\node[align=center] at (3,-2.5) {\Large{Machine 3}};
\draw (4,-2) [fill=PastelPurple]rectangle ++(1,-1) node[midway](M3){0.6};
\draw (5,-2) [fill=PastelBlue]rectangle ++(1,-1) node[midway]{1};
\draw (6,-2) [fill=PastelOrange]rectangle ++(1,-1) node[midway]{0.1};
\node[anchor=west] at (7,-2.5) {Machine 3};

%Machine Descriptions
\node[align=center] at (4.5,-3.5) {$r_m^{(s)}$};
\node[align=center] at (5.5,-3.5) {$\epsilon_m^{(s)}$};
\node[align=center] at (6.5,-3.6) {$\omega_m$}; %-3.5

%Lines and Arrows
\draw[->, shorten >=15pt, shorten <=15pt] (M1) -- (P1) node[above, pos=0.5] {\Large{$x_1^{(j)}$}};
\draw[->, shorten >=15pt, shorten <=15pt] (M2) -- (P2) node [above, pos=0.5] {\Large{$x_2^{(j)}$}};
\draw[->, shorten >=15pt, shorten <=15pt] (M3) -- (P3) node [above, pos=0.5] {\Large{$x_3^{(j)}$}};

\draw[-, red, ultra thick] (-1,-3) -- (2,-3);
\draw [decorate, decoration={brace, raise= 3pt, amplitude=10pt}] (1,-3.05) --  (1,-4.95);
\node[align=center] at (2,-4) {\Large{$x_{\text{urg}}^{(j)}$}};

\draw [decorate, decoration={brace, raise= 3pt, amplitude=20pt}] (0,0) --  (7,0);
\node[align=center] at (3.5,1.25) {\Large{$X_{res}^{(j)}$}};

\end{tikzpicture}}}
        \caption{Data to input: $X^{(j)}_{\text{res}}$ and $x^{(j)}_{\text{urg}}$}
        \label{fig:data to input}
    \end{subfigure}
    \hfill
    %\vspace{15pt}
    \begin{subfigure}[b]{0.5\textwidth}%{0.43\textwidth}
        \centering
        \resizebox{\textwidth}{!}{%
{\scalefont{1.5}%
\begin{tikzpicture}

%Columns Legend
\node[align=center] at (0.5,2.5) {$p_{jm}$};
\node[align=center] at (1.5,2.6) {$r_m^{(s)}$}; %2.5
\node[align=center] at (2.5,2.6) {$\epsilon_m^{(s)}$}; %2.5
\node[align=center] at (3.5,2.5) {$\omega_m$};

%Input
	%M1
\node[align=center] at (-0.5,1.5) {\Large{$x_1^{(j)}$}};
\draw (0,2) [fill=Green1]rectangle ++(1,-1) node[midway]{0.3};
\draw (1,2) [fill=PastelPurple]rectangle ++(1,-1) node[midway]{0};
\draw (2,2) [fill=PastelBlue]rectangle ++(1,-1) node[midway]{0.4};
\draw (3,2) [fill=PastelOrange]rectangle ++(1,-1) node[midway]{0.7};

	%M2
\node[align=center] at (-0.5,-0.5) {\Large{$x_2^{(j)}$}};
\draw (0,-0) [fill=Green2]rectangle ++(1,-1) node[midway]{0.9};
\draw (1,-0) [fill=PastelPurple]rectangle ++(1,-1) node[midway]{0.2};
\draw (2,-0) [fill=PastelBlue]rectangle ++(1,-1) node[midway]{0.5};
\draw (3,-0) [fill=PastelOrange]rectangle ++(1,-1) node[midway]{1};

	%M3
\node[align=center] at (-0.5,-2.5) {\Large{$x_3^{(j)}$}};
\draw (0,-2) [fill=Green3]rectangle ++(1,-1) node[midway]{0.5};
\draw (1,-2) [fill=PastelPurple]rectangle ++(1,-1) node[midway]{0.6};
\draw (2,-2) [fill=PastelBlue]rectangle ++(1,-1) node[midway]{1};
\draw (3,-2) [fill=PastelOrange]rectangle ++(1,-1) node[midway]{0.1};

%NN
\draw (6,0.5) [fill=LightGray]rectangle ++(2,-2);
\node[align=center] at (7,-.7) {LSTM\\bidirect.};

%Output
\node[align=center] at (10.6,1.7) {static size\\Embedding};
\draw (10,1) [fill=CarnationPink]rectangle ++(1.2,-1.5) node[midway]{$\overrightarrow{x}_{\text{res}}^{(j)}$};
\draw (10,-0.5) [fill=CarnationPink]rectangle ++(1.2,-1.5) node[midway]{$\overleftarrow{x}_{\text{res}}^{(j)}$};

%Final Output
\draw [decorate, decoration={brace, raise= 3pt, amplitude=10pt}] (11.2,1-0.05) --  (11.2,-1.95);
\node[align=center] at (12.2,-0.5) {\Large{$x_{\text{res}}^{(j)}$}};

%Arrows
\draw[->, shorten >= 15pt] (4,1.5) -- (6,-0.5);
\draw[->, shorten >= 15pt] (4,-0.5) -- (6,-0.5);
\draw[->, shorten >= 15pt] (4,-2.5) -- (6,-0.5);

\draw[->, shorten >= 10pt] (8,-0.5) -- (10,1-0.75);
\draw[->, shorten >= 10pt] (8,-0.5) -- (10,-1.25);

\end{tikzpicture}}}
\caption{Sequential embedding of $X_{\text{res}}^{(j)}$ into fixed-sized vector}
\label{fig:input to LSTM}
    \end{subfigure}
    \caption{Embedding machine data into job representations of static dimension}
    \label{fig:data embedding}
\end{figure*}

Our model processes each emerging state $s \in S$ with $J$ pending jobs and $M$ active machines to induce scheduling actions. The sorted state data is the input for our neural network. The network's output $\hat{y} \in \mathbb{R}^{J+1}$ is a probability distribution over the feasible actions: $J$ for assigning each job to the available machine $m$ and one additional action for deactivating the machine. The architecture comprises three main components, of which the first two heavily take advantage of parallelization:
\begin{enumerate}
\item \textbf{Sequential Embedding}: Generates fixed-size representation for each job by processing and embedding machines data pertaining to it as time series via an LSTM.
\item \textbf{Transformer Encoder}: Enhances information of job representations by integrating inter-jobs context.
\item \textbf{Action-Pointer Decoder}: Produces probability distribution across all feasible actions, leveraging comprehensive context of jobs and machines.
\end{enumerate}
Dense feedforward layers (FF) with Leaky ReLU as activation function $\phi_F$ and activation $\alpha=0.1$ are used for further enhancements and can be computed in parallel as well:
\begin{equation}
FF(x) := W_{\text{out}}*\phi_F(W_{\text{in}}*x+b_{\text{in}})+b_{\text{out}}.
\label{eq:FF}
\end{equation} 

\subsection{Data Preprocessing}
\label{section:data_preprocessing}

As detailed in \cref{section:states_as_inputs}, each state $s \in S$ features a variable input size, including $M$ active machine vectors of dimension 3 and $J$ pending job vectors, whose feature dimensions $M+2$ furthermore vary with the number of machines, complicating the standardization of inputs for neural network processing. To address this, we define the vector
\begin{equation}
    x_m^{(j)} := [p_{jm}, r_m^{(s)}, \epsilon_m^{(s)}, \omega_m]^T \in \mathbb{R}^4
\end{equation}
representing all information of a machine $m$ with respect to a job $j$ and subsequently decompose each job's data into two distinct components:
\begin{align}
    X_{\text{res}}^{(j)} &:= [x_1^{(j)}, \ldots, x_M^{(j)}]^T \in \mathbb{R}^{M \times 4}\\
    x_{\text{urg}}^{(j)} &:= [e_j^{(s)}, w_j]^T \in \mathbb{R}^2.
\end{align}
For each job, the first component embeds the data of every machine related to the job into its representation. This is achieved by organizing the machine data as a time series matrix, representing the emerging resources for the job. The second component is fixed in size and captures the urgency of processing each job. Note that $J$ and $M$ refer to \emph{pending} jobs and \emph{active} machines and hence vary throughout the scheduling process. To ensure consistent input scales, weights are normalized by being divided by the maximum allowed value, while all time-related data is normalized by the highest time entry in each state. For a visual example of this data transformation process, see \cref{fig:data to input}. This dual-part approach allows our model to generate fixed-size job vectors including the processed machine information, as shown in \cref{fig:data embedding} and detailed in the next \cref{section:sequential_embedding}.
For all jobs, the neural network thus receives a resource tensor and an urgency matrix as normalized inputs:
\begin{align}
    \text{Resource Input : } &\left[ X_{\text{res}}^{(1)}, \ldots, X_{\text{res}}^{(J)} \right]^T \in [0,1]^{J \times M \times 4} \label{eq:res_input} \\
    \text{Urgency Input : } &\left[ x_{\text{urg}}^{(1)}, \ldots, x_{\text{urg}}^{(J)} \right]^T \in [0,1]^{J \times 2}. \label{eq:urg_input} 
\end{align}

\subsection{Sequential Embedding} \label{section:sequential_embedding}

The input of the network consists of the two lists of job representations defined in (\ref{eq:res_input}) and (\ref{eq:urg_input}). For every job $j$, its time series of emerging machine resources ${X_{\text{res}}^{(j)} = x_1^{(j)}, \ldots, x_M^{(j)}}$ from (\ref{eq:res_input}) depends on the number of machines $M$. In parallel, each $X_{\text{res}}^{(j)}$ is embedded into a fixed-size representation $x_{\text{res}}^{(j)}$ by being passed through a bidirectional LSTM, as illustrated in \cref{fig:input to LSTM}. $x_{\text{res}}^{(j)}$ is then refined to $\tilde{x}_{\text{res}}^{(j)}$ using an FF layer.

\begin{figure}
\centering
%\resizebox{0.7\textwidth}{!}{%
\resizebox{0.47\textwidth}{!}{%
{\scalefont{1.4}%
\begin{tikzpicture}
	
	%Sizes
\def\xd{6}
\def\yd{-2}
\def\r{0.6}

	%Inputs
\draw (0*\xd,-0.5) [fill=PastelGreen] circle (\r cm) node(x1){$x_{\text{urg}}^{(1)}$};
\node [align=center] at (0*\xd,\yd+0.1) (x_dots){\Huge{$\vdots$}};
\draw (0*\xd,2*\yd + 0.5) [fill=PastelGreen] circle (\r cm) node(xJ){$x_{\text{urg}}^{(J)}$};

	%Self Attention NN
\draw (0.7*\xd,0.5*\yd)[fill=PastelBlue] rectangle ++ (0.6*\xd,\yd) node[midway](hT_1){\Large{Self-Attention}};

	%Outputs
\draw (2*\xd,-0.5) [fill=PastelOrange] circle (\r cm) node(z1){$\tilde{x}_{\text{urg}}^{(1)}$};
\node [align=center] at (2*\xd,\yd+0.1) (x_dots){\Huge{$\vdots$}};
\draw (2*\xd,2*\yd + 0.5) [fill=PastelOrange] circle (\r cm) node(zJ){$\tilde{x}_{\text{urg}}^{(J)}$};

	%Edges
	
%Inputs to NN
\draw[->, shorten <>=8pt] (x1) -- (0.7*\xd,1*\yd) node[midway, above=3pt]{$FF_{\text{urg}}^{(1)}$};
\draw[->, shorten <>=8pt] (xJ) -- (0.7*\xd,1*\yd) node[midway, below=5pt]{$FF_{\text{urg}}^{(1)}$};

%NN to Outputs
\draw[->, shorten <>=8pt] (1.3*\xd,1*\yd) -- (z1) node[midway, above=5pt]{$FF_{\text{urg}}^{(2)}$};
\draw[->, shorten <>=8pt] (1.3*\xd,1*\yd) -- (zJ) node[midway, below=6pt]{$FF_{\text{urg}}^{(2)}$};

\end{tikzpicture}}}
\caption{Contextualizing urgencies of jobs}
\label{fig:context_urgency}
%\vspace{6pt}
\end{figure}
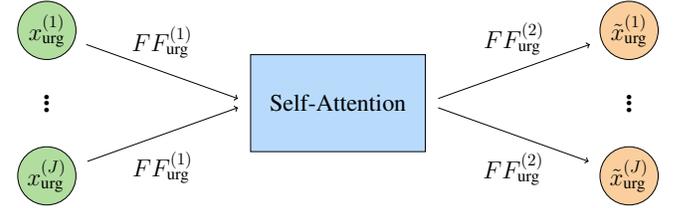

Also in parallel, the inputs $x_{\text{urg}}^{(j)}$ from (\ref{eq:urg_input}) are processed to contextualize each job’s urgency relative to others, as illustrated in \cref{fig:context_urgency}. This is achieved by first passing the vectors through an FF layer, then applying a self-attention mechanism to produce contextual urgency representations, and finally refining the output with another FF layer to obtain $\tilde{x}_{\text{urg}}^{(j)}$.

The final embedded representation $x_{\text{emb}}^{(j)}$ for each job is obtained by concatenating $\tilde{x}_{\text{res}}^{(j)}$ and $\tilde{x}_{\text{urg}}^{(j)}$ and passing them through an additional FF layer to further catch context.  The resulting sequence of all embedded job representations is:
\begin{equation}
X_{\text{emb}} := [x_{\text{emb}}^{(1)}, \ldots, x_{\text{emb}}^{(J)}]^T \in \mathbb{R}^{J \times 16}.
\label{eq:init_embed}
\end{equation}

\subsection{Transformer Encoder}

Each job is initially embedded with context regarding its urgency and specific machine data. However, it lacks knowledge about the processing times of other jobs on the active machines. To integrate this information, a Transformer encoder is applied to (\ref{eq:init_embed}), as described by~\cite{vaswani2017attention}. This encoder utilizes a self-attention mechanism, effectively broadening the context of $x_{\text{emb}}^{(j)}$ to include how other jobs interact with the same machines. The output of this encoder, $z^{(j)} \in \mathbb{R}^{16}$, provides a comprehensive encoding of job $j$ within the full context of the current state of the given scheduling problem.

\subsection{Action-Pointer Decoder}

\tikzset{embed_vec/.style = {matrix of math nodes,left delimiter=(,right delimiter=)},
shorten <>/.style={ shorten >=#1, shorten <=#1 }}

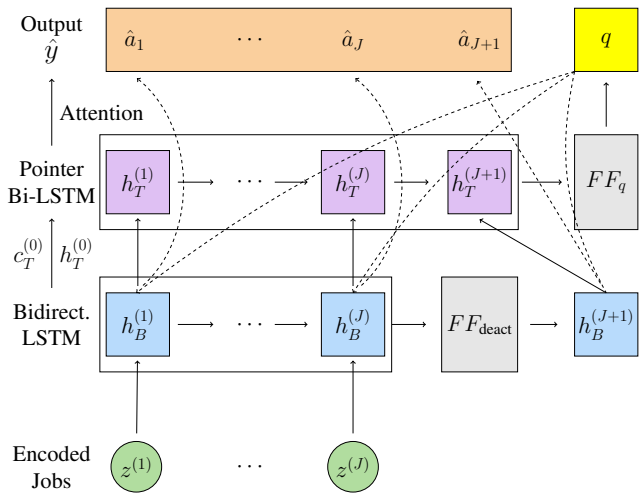
\begin{figure}[t]
\centering
\resizebox{0.47\textwidth}{!}{%
%\resizebox{0.9\textwidth}{!}{%
{\scalefont{2}%
\begin{tikzpicture}%[yscale=1.05]
	%Sizes
\def\xd{4}
\def\yd{4.5}
\def\r{0.8}

	%Titles
\node[align=center] at (-0.38*\xd,0*\yd) (y) {\huge{Output}\\\Huge{$\hat{y}$}};
\node[align=center] at (-0.38*\xd,-1*\yd) (T-LSTM) {\huge{Pointer}\\\huge{Bi-LSTM}};
\node[align=center] at (-0.38*\xd,-2*\yd) (B-LSTM) {\huge{Bidirect.}\\\huge{LSTM}};
\node[align=center] at (-0.38*\xd,-3*\yd) 
{\huge{Encoded}\\\huge{Jobs}}; %Inputs

	%Nodes
%Outputs	
\draw (0.3*\xd-1,0*\yd+1) [fill=PastelOrange] rectangle ++(2.7*\xd+2,-2);
	
%Output
\node at (1.1,0*\yd) (a_1){\huge{$\hat{a}_1$}};
\node at (1.2*\xd,0*\yd) {\huge{$\cdots$}};
\node at (2*\xd,0*\yd) (a_J){\huge{$\hat{a}_J$}};
\node at (3*\xd,0*\yd) (a_J1){\huge{$\hat{a}_{J+1}$}};
\draw (4*\xd-1,0*\yd+1)[fill=Yellow] rectangle ++ (2,-2) node[midway](q){\huge{$q$}};

%Top LSTM
\draw (0*\xd,-1*\yd+1.5) rectangle ++(3.305*\xd,-3);
\draw (0.3*\xd-1,-1*\yd+1)[fill=PastelPurple] rectangle ++ (2,-2) node[midway](hT_1){$h_T^{(1)}$}; 
\node[align=center] at (1.2*\xd,-1*\yd) (hT_dots){\huge{$\cdots$}};
\draw (2*\xd-1,-1*\yd+1)[fill=PastelPurple] rectangle ++ (2,-2) node[midway](hT_J){$h_T^{(J)}$};
\draw (3*\xd-1,-1*\yd+1)[fill=PastelPurple] rectangle ++ (2,-2) node[midway](hT_J1){$h_T^{(J+1)}$};
\draw (4*\xd-1,-1*\yd+1.5)[fill=LightGray] rectangle ++ (2,-3) node[midway](FF2){$FF_{q}$};

%Bottom LSTM
\draw (0*\xd,-2*\yd+1.5) rectangle ++(2.305*\xd,-3);
\draw (0.3*\xd-1,-2*\yd+1)[fill=PastelBlue] rectangle ++ (2,-2) node[midway](hB_1){$h_B^{(1)}$};
\node at (1.2*\xd,-2*\yd) (hB_dots){\huge{$\cdots$}};
\draw (2*\xd-1,-2*\yd+1)[fill=PastelBlue] rectangle ++ (2,-2) node[midway](hB_J){$h_B^{(J)}$};
\draw (3*\xd-1.2,-2*\yd+1.5)[fill=LightGray] rectangle ++ (2.4,-3) node[midway](FF1){$FF_{\text{deact}}$};
\draw (4*\xd-1,-2*\yd+1)[fill=PastelBlue] rectangle ++ (2,-2) node[midway](hB_J1){$h_B^{(J+1)}$};

%Input
\draw (1.1525,-3*\yd) [fill=PastelGreen] circle (\r cm) node(z_1){$z^{(1)}$};
\node at (1.2*\xd,-3*\yd) (z_dots){\huge{$\cdots$}};
\draw (2*\xd,-3*\yd) [fill=PastelGreen] circle (\r cm) node(z_J){$z^{(J)}$};

	%Edges

\draw[->, thick, shorten <>=7pt] (T-LSTM) -- (y) node[right, pos=0.5] {\huge{Attention}};
\draw[->, thick, shorten <>=7pt] (B-LSTM) -- (T-LSTM) node[right, pos=0.5] {$h_T^{(0)}$} node[left, pos=0.5] {$c_T^{(0)}$};

%Output
\draw[dashed, thick] (4*\xd-1,0*\yd-1) edge[->, bend right=10] (1.15,-2*\yd+1);
\draw[dashed, thick] (4*\xd-1,0*\yd-1) edge[->, bend right=12] (2*\xd,-2*\yd+1);
\draw[dashed, thick] (4*\xd-1,0*\yd-1) edge[-, bend right=20] (4*\xd,-2*\yd+1);

\draw [, dashed, thick, shorten <>=10pt] (1.1,-2*\yd+1) edge[->, bend right=50] (0.9,0*\yd-1);
\draw [, dashed, thick, shorten <>=10pt] (2*\xd,-2*\yd+1) edge[->, bend right=50] (2*\xd-0.2,0*\yd-1);
\draw[->, dashed, thick, shorten <>=10pt] (4*\xd,-2*\yd+1) -- (3*\xd-0.25,0*\yd-1);

%\draw[->, dashed, thick, shorten <=15pt, shorten >=25pt] (hT_1) -- (a_1);
%\draw[->, dashed, thick, shorten <=15pt, shorten >=25pt] (hT_J) -- (a_J);
%\draw[->, dashed, thick, shorten <=15pt, shorten >=25pt] (hT_J1) -- (a_J1);

%Top LSTM
\draw[->, thick, shorten <>=14pt] (hT_1) -- (hT_dots);
\draw[->, thick, shorten >=18pt] (hT_dots) -- (hT_J);
\draw[->, thick, shorten <>=14pt] (hT_J) -- (hT_J1);
\draw[->, thick, shorten <=6pt, shorten >=15pt] (hT_J1) -- (FF2);

\draw[->, thick, shorten <=33pt, shorten >=25pt] (FF2) -- (q);

%Bottom LSTM
\draw[->, thick, shorten <>=14pt] (hB_1) -- (hB_dots);
\draw[->, thick, shorten >=18pt] (hB_dots) -- (hB_J);
\draw[->, thick, shorten <=12pt, shorten >=12pt] (hB_J) -- (FF1);
\draw[->, thick, shorten <=9pt, shorten >=12pt] (FF1) -- (hB_J1);

\draw[->, thick, shorten <=14pt, shorten >=12pt] (hB_1) -- (hT_1);
\draw[->, thick, shorten <=14pt, shorten >=12pt] (hB_J) -- (hT_J);
\draw[->, thick, shorten <=0pt] (4*\xd,-2*\yd+1) -- (3*\xd,-1*\yd-1.18); %hB_J1.north east 32pt
%Inputs
\draw[->, thick, shorten <=14pt, shorten >=12pt] (z_1) -- (hB_1);
\draw[->, thick, shorten <=14pt, shorten >=12pt] (z_J) -- (hB_J);

\end{tikzpicture}}}

%\caption{Action-pointer decoder}

\caption{The modified Action-Pointer decoder outputs an attention distribution over all feasible actions: assigning the associated pending job to the currently free machine ($\hat{a}_1, \ldots, \hat{a}_J$) or deactivating the machine ($\hat{a}_{J+1}$).}

%y\caption{Modified action-pointer decoder with integrated dense feedforward layers $FF$. Encoded job representations with immanent machines context are processed by the bottom LSTM to generate an additional representation for deactivating the currently unoccupied machine. The resulting $J+1$ vectors are then passed as keys to the top LSTM to generate a query~$q$. The output is an attention distribution over all feasible actions.}

%\caption{Decoder. The neural network employs a modified action-pointer as decoder with integrated feedforward layers $FF$. Job representations embedded into the machines context are provided by a Transformer encoder. These are processed by the bottom LSTM to generate an additional representation for deactivating the currently unoccupied machine. The resulting $J+1$ vectors are then passed as keys to the top LSTM to generate a query~$q$. The output is an attention distribution over all feasible actions.}
\label{fig:Pointer Decoder}
\end{figure}

The output of our network aims to include one entry for each of the $J$ pending jobs plus an additional entry for the option to deactivate the machine, thus necessitating a variable-dimensional output. We employ an Action-Pointer decoder~\cite{vinyals2015pointer} to achieve this flexibility effectively. \cref{fig:Pointer Decoder} illustrates the complete architecture of the decoder.

First, the encoded job representations $z^{(1)}, \ldots, z^{(J)}$ are processed through a bidirectional LSTM (bottom LSTM), generating a sequence of hidden states $h_B^{(1)}, \ldots, h_B^{(J)} \in \mathbb{R}^{64}$, each correlating to the potential action of assigning the corresponding job to the currently free machine. To create a context-aware representation for the deactivation action~$h_B^{(J+1)}$, we pass the final hidden state $h_B^{(J)}$ through an FF layer.
Subsequently, both the final hidden and cell states of the bottom LSTM are processed through another FF layer to initialize the memory states of a second bidirectional LSTM (top LSTM), effectively forming our action-pointer. It sequentially processes the hidden states from the bottom LSTM. These also serve as keys. Another FF layer is applied to its final hidden state $h_T^{(J+1)}$ to produce the query $q \in \mathbb{R}^{128}$. The neural network’s final output is then an attention distribution, computed by applying the attention mechanism described in~\cite{vinyals2015pointer} with softmax activation to normalize the scores across the $J+1$ actions:
\begin{equation*}
\hat{y}=[\hat{a}_1, \ldots, \hat{a}_{J+1}]^T \in [0,1]^{J+1}.   
\end{equation*}
\begin{table}[t]
    \centering
    \caption{Network parameters and dimensions\\}
    %\vspace{5pt}
    \label{tab:network_dims}
    \resizebox{0.45\textwidth}{!}{%
    \renewcommand{\arraystretch}{1.12}
    \begin{tabular}{lcc}
    \toprule
Layer & \begin{tabular}[c]{@{}c@{}}Parameters\\ (total = 117.292)\end{tabular} & \begin{tabular}[c]{@{}l@{}}Output\\ Dimension\end{tabular} \\ \midrule
Bidirectional LSTM & 2688 & $2 \cdot 16$\\
$FF_{\text{res}}$ & 800 & 16\\
$FF_{\text{urg}}^{(1)}$ & 32 & 4\\
Self-Attention & 308 & 4\\
$FF_{\text{urg}}^{(2)}$ & 40 & 4\\
Concatenate Inputs & 0 & 20\\
$FF_{\text{emb}}$ & 608 & 16\\ \midrule
Sequential Embedding & total = 4476 & \begin{tabular}[c]{@{}l@{}}Input : $J \times M \times 4 + J \times 2$\\ Output : $J \times 16$\end{tabular}\\ \midrule

Self-Attention Encoder & 8592 & 16\\
Residual Connection 1 & 0 & 16\\
$FF_{\text{enc}}$ & 544 & 16\\
Residual Connection 2 & 0 & 16 \\ \midrule
Transformer Encoder & total = 9136 & \begin{tabular}[c]{@{}l@{}}Input : $J \times 16$\\ Output : $J \times 16$\end{tabular} \\ \midrule

\begin{tabular}[c]{@{}l@{}}Decoder Bottom\\ Bidirectional LSTM\end{tabular} & 12544 & $J \times (2\cdot 32)$\\
$FF_{\text{deact}}$  & 8320 & 64\\
\begin{tabular}[c]{@{}l@{}}Concatenate $h_B^{(J+1)}$\\ to Sequence\end{tabular} & 0 & $(J+1)\times 64$\\
$FF_{\overrightarrow{h}}$ & 2112 & 32\\
$FF_{\overleftarrow{h}}$ & 2112 & 32\\
$FF_{\overrightarrow{c}}$ & 2112 & 32\\
$FF_{\overleftarrow{c}}$ & 2112 & 32\\
\begin{tabular}[c]{@{}l@{}}Action-Pointer:\\ Decoder Top\\ Bidirectional LSTM\end{tabular} & 24832 & $(J+1) \times (2 \cdot 32)$\\
$FF_{q}$ & 24832 & 128\\
\begin{tabular}[c]{@{}l@{}}Action-Pointer:\\ Cross-Attention\end{tabular} & 24704 & $J+1$\\ \midrule 
Action-Pointer & total = 103.680 & \begin{tabular}[c]{@{}l@{}}Input : $J \times 16$\\ Output : $J + 1$\end{tabular} \\ \bottomrule
\end{tabular}}
\end{table}
An overview over all network parameters and dimensions is given by \cref{tab:network_dims}.
\section{Empirical Evaluation}

In addition to the 100.000 scheduling problems used for training, we have simulated 10.000 problems in the form of (\ref{our_JSP}) and according to the constraints of \cref{section:data} for validation, and another 10.000 for testing.
From every such problem, one state has been randomly selected for each combination of 3 to 8 pending jobs and 2 to 4 active machines. To evaluate our neural network, we employed the metric $\Delta Q(s,\mu_\vartheta(s))$ defined in (\ref{eq:training_metric}), which measures the additional costs incurred by following the policy $\mu_\vartheta$, induced by our network with learned parameters $\vartheta$, compared to the optimal policy $\mu^*$. The results, presented in \cref{table:sup_learn_res}, demonstrate that our model consistently achieves additional relative costs  averaging below 1\% across all settings, underscoring its robust generalization capabilities. Notably, performance even tended to be better during validation and testing. This is attributed to the balanced selection of states during training, while the random selection during testing likely included a higher proportion of states where our initial sorting permutation $\Pi_m$ already decently approximated $\mu^*$, thus simplifying scheduling optimization.

\subsection{Supervised Performance for 8 Jobs and 4 Machines}

\begin{table*}[t]
    \centering
    \caption{Relative increase to optimal costs when scheduling 8 jobs and 4 machines}
    \begin{subtable}[t]{0.45\textwidth}
        \centering
        \renewcommand{\arraystretch}{0.5}
        \caption{Per state decision in schedule}
        \resizebox{\textwidth}{!}{%
{\scalefont{1.2}%
\begin{tabular}{ccccc}
\#Jobs & \#Machines & \begin{tabular}[c]{@{}c@{}}Training\\ $\Delta Q$ [\%]\end{tabular} & \begin{tabular}[c]{@{}c@{}}Validation\\ $\Delta Q$ [\%]\end{tabular} & \begin{tabular}[c]{@{}c@{}}Test\\ $\Delta Q$ [\%]\end{tabular} \\ \midrule%\hline \hline
 & 2 & $0.45$ & $0.14$ & $0.15$ \\
3 & 3 & $0.33$ & $0.26$ & $0.23$ \\
 & 4 & $0.36$ & $0.46$ & $0.45$ \\
\midrule%\hline
 & 2 & $0.63$ & $0.16$ & $0.13$ \\
4 & 3 & 0.51 & $0.15$ & $0.21$ \\
 & 4 & $0.24$ & $0.31$ & $0.32$ \\
\midrule%\hline
 & 2 & $0.58$ & $0.17$ & $0.17$ \\
5 & 3 & $0.60$ & $0.22$ & $0.24$ \\
 & 4 & $0.53$ & $0.31$ & $0.31$ \\
\midrule%\hline
 & 2 &  $0.28$ & $0.20$ & $0.21$ \\
6 & 3 & $0.47$ & $0.28$ & $0.32$ \\
 & 4 & $0.45$ & $0.32$ & $0.35$ \\
\midrule%\hline
 & 2 & $0.31$ & $0.20$ & $0.19$ \\
7 & 3 & $0.45$ & $0.40$ & $0.43$ \\
 & 4 & $0.45$ & $0.41$ & $0.40$ \\
\midrule%\hline
 & 2 & $0.34$ & $0.32$ & $0.32$ \\
8 & 3 & $0.30$ & $0.60$ & $0.57$ \\
 & 4 & $0.45$ & $0.63$ & $0.63$     
\end{tabular}}}
        \label{table:sup_learn_res}
    \end{subtable}
    \hfill
    %\vspace{15pt}
    \begin{subtable}[t]{0.42\textwidth}
        \centering
        \renewcommand{\arraystretch}{2}
        \caption{For the total schedule}
\resizebox{\textwidth}{!}{%
{\scalefont{0.8}%
\begin{tabular}{ccc}
Method                                                                       & Policy & $\Delta V_\mu$ [\%] \\ \hline
\begin{tabular}[c]{@{}l@{}}Neural Network\\ early optimal cutoff\end{tabular} & $\mu_{\vartheta,\text{opt}}$ & $2.51$ \\ \hline
Neural Network & $\mu_\vartheta$ & $4.47$ \\ \hline
\begin{tabular}[c]{@{}l@{}}Dispatching Rule\\ early optimal cutoff\end{tabular} & $\mu_{\text{rule,opt}}$ & $34.46$\\ \hline
Dispatching Rule & $\mu_{\text{rule}}$ & $45.33$
\end{tabular}}}

        \label{table:sup_pol_results}

    \end{subtable}
    \vspace{10pt}
    \end{table*}
\definecolor{PastelGreen2}{RGB}{67,185,09}
\definecolor{PastelBlue}{RGB}{184,219,252}
\definecolor{PastelOrange}{RGB}{250,207,157}
\definecolor{LightGray}{gray}{0.9}

\begin{table*}[t]
\centering
\vspace{-1pt}
\caption{Relative increase [\%] in scheduling costs for using dispatching rule over network for different numbers of jobs ($J$) and machines ($M$). Values are defined by (\ref{eq:rule_over_network}), given as percentages and represent the performance advantage of our network.}
%\vspace{-8pt}
%\renewcommand{\arraystretch}{0.5}
\resizebox{0.9\textwidth}{!}{%
{\scalefont{0.9}%
\begin{tabular}{|c||*{7}{c}c|}\hline
\rowcolor{PastelBlue!30}\scalebox{0.76}{\backslashbox{{\large J}}{\raisebox{-1.2ex}{\large M}}}%{J \textbackslash \ M}\backslashbox{J}{M}
&\makebox[1em]{2}&\makebox[1em]{3}&\makebox[1em]{4}
&\makebox[1em]{5}&\makebox[1em]{6}&\makebox[1em]{8}&\makebox[1em]{10}&\makebox[1em]{12}\\\hline\hline

\cellcolor{PastelBlue!30}8
&\cellcolor{PastelGreen2!48.925}19.57&\cellcolor{PastelGreen2!68.675}27.47&\cellcolor{PastelGreen2!77.925}31.17&\cellcolor{PastelGreen2!80.225}32.09&\cellcolor{PastelGreen2!74.475}29.79&\cellcolor{PastelGreen2!45.325}18.13&\cellcolor{PastelGreen2!20.8}08.32&\cellcolor{PastelGreen2!3.675}01.47\\

\cellcolor{PastelBlue!30}9
&\cellcolor{PastelGreen2!52.775}21.11&\cellcolor{PastelGreen2!63.45}25.38&\cellcolor{PastelGreen2!80.975}32.39&\cellcolor{PastelGreen2!84.475}33.79&\cellcolor{PastelGreen2!81.1}32.44&\cellcolor{PastelGreen2!50.55}20.22&\cellcolor{PastelGreen2!23.45}09.38&\cellcolor{red!1.44} -01.44\\

\cellcolor{PastelBlue!30}10
&\cellcolor{PastelGreen2!50.225}20.09&\cellcolor{PastelGreen2!64.05}25.62&\cellcolor{PastelGreen2!80.65}32.26&\cellcolor{PastelGreen2!88.575}35.43&\cellcolor{PastelGreen2!86.075}34.43&\cellcolor{PastelGreen2!57.55}23.02&\cellcolor{PastelGreen2!27.725}11.09&\cellcolor{PastelGreen2!0.725}00.29\\

\cellcolor{PastelBlue!30}11
&\cellcolor{PastelGreen2!45.05}18.02&\cellcolor{PastelGreen2!67.15}26.86&\cellcolor{PastelGreen2!82.575}33.03&\cellcolor{PastelGreen2!90.525}36.21&\cellcolor{PastelGreen2!89.7}35.88&\cellcolor{PastelGreen2!63.05}25.22&\cellcolor{PastelGreen2!29.125}11.65&\cellcolor{PastelGreen2!5.1}02.04\\

\cellcolor{PastelBlue!30}12
&\cellcolor{PastelGreen2!45.725}18.83&\cellcolor{PastelGreen2!65.825}26.33&\cellcolor{PastelGreen2!81.075}32.43&\cellcolor{PastelGreen2!91.325}36.53&\cellcolor{PastelGreen2!92.725}37.09&\cellcolor{PastelGreen2!65.9}26.36&\cellcolor{PastelGreen2!31.525}12.61&\cellcolor{PastelGreen2!7.25}02.90\\

\cellcolor{PastelBlue!30}15
&\cellcolor{PastelGreen2!44.95}17.98&\cellcolor{PastelGreen2!55.18}22.07&\cellcolor{PastelGreen2!75.75}30.30&\cellcolor{PastelGreen2!92.48}
36.99&\cellcolor{PastelGreen2!95.67}38.27&\cellcolor{PastelGreen2!73.10}29.24&\cellcolor{PastelGreen2!37.93}15.17&\cellcolor{PastelGreen2!8.70}03.48\\

\cellcolor{PastelBlue!30}20
&\cellcolor{PastelGreen2!37.85}15.14&\cellcolor{PastelGreen2!53.23}21.29&\cellcolor{PastelGreen2!67.25}26.90&\cellcolor{PastelGreen2!85.57}34.23&\cellcolor{PastelGreen2!95.35}38.14&\cellcolor{PastelGreen2!78.82}31.53&\cellcolor{PastelGreen2!42.72}17.09&\cellcolor{PastelGreen2!12.18}04.87\\

\cellcolor{PastelBlue!30}30
&\cellcolor{PastelGreen2!33.40}13.36&\cellcolor{PastelGreen2!41.83}16.73&\cellcolor{PastelGreen2!53.00}21.20&\cellcolor{PastelGreen2!69.35}27.74&\cellcolor{PastelGreen2!81.55}32.62&\cellcolor{PastelGreen2!79.30}31.72&\cellcolor{PastelGreen2!45.10}18.04&\cellcolor{PastelGreen2!10.98}04.39\\

\cellcolor{PastelBlue!30}50
&\cellcolor{PastelGreen2!26.625}10.65&\cellcolor{PastelGreen2!31.3}12.52&\cellcolor{PastelGreen2!37.35}14.94&\cellcolor{PastelGreen2!46.4}18.56&\cellcolor{PastelGreen2!58.5}23.40&\cellcolor{PastelGreen2!65.325}26.13&\cellcolor{PastelGreen2!40.9}16.36&\cellcolor{PastelGreen2!10.325}04.13\\

\cellcolor{PastelBlue!30}75
&\cellcolor{PastelGreen2!24.775}09.91&\cellcolor{PastelGreen2!25.225}10.09&\cellcolor{PastelGreen2!26.40}10.56&\cellcolor{PastelGreen2!31.725}12.69&\cellcolor{PastelGreen2!37.425}14.97&\cellcolor{PastelGreen2!45.05}18.02&\cellcolor{PastelGreen2!31.07}12.68&\cellcolor{PastelGreen2!9.3}03.72\\

\cellcolor{PastelBlue!30}100
&\cellcolor{PastelGreen2!24.845}09.94&\cellcolor{PastelGreen2!22.40}08.96&\cellcolor{PastelGreen2!19.925}07.97&\cellcolor{PastelGreen2!18.975}07.59&\cellcolor{PastelGreen2!21.125}08.45&\cellcolor{PastelGreen2!27.875}11.15&\cellcolor{PastelGreen2!22.75}09.10&\cellcolor{PastelGreen2!8.175}03.27\\\hline
\end{tabular}}}
\label{table:generalized_results}
\end{table*}

We assessed our neural network’s performance by generating complete schedules for a new set of 10,000 scheduling problems with 8 jobs and 4 machines. The costs incurred using our network’s policy $\mu_\vartheta$ were compared to the optimal costs and against those produced by the heuristic dispatching rule from \cref{section:heuristic}, denoted by $\mu_{\text{rule}}$. To focus on more complex decision-making scenarios, we defined a second version of each policy, denoted by the addition \emph{opt}, where the remaining optimal schedule is directly computed when only two jobs remain or only one machine is active with not more than 8 pending jobs (important for the next \cref{section:higher_problems}). For each scheduling problem represented by an initial state $s$, we measured the relative increase in scheduling costs compared to the optimal costs $V^*(s)$ with the metric:
\begin{equation}
\Delta V_\mu(s) := \frac{V_\mu(s)-V^*(s)}{V^*(s)}.
\label{eq:additional_costs_compared_to_opt}
\end{equation}
Results are summarized in \cref{table:sup_pol_results}, validating that our network closely approximates the optimal scheduling costs. Moreover, the network strongly outperforms the dispatching rule, inducing additional relative costs that are less than 10\% of those produced by the heuristic algorithm.

\subsection{Performance for Higher Scheduling Problems}\label{section:higher_problems}

We evaluate how well our neural network generalizes for larger and more complex scheduling problems compared to the dispatching rule. To this end, we have simulated 10.000 additional scheduling problems with constraints as outlined in \cref{section:data} for each of the various combinations of job numbers $J$ and machine numbers $M$ presented in \cref{table:generalized_results}. The entries of this table represent the quotient
\begin{equation}
\frac{V_{\mu_{\text{rule,opt}}}(s)-V_{\mu_{\vartheta,\text{opt}}}(s)}{V_{\mu_{\vartheta,\text{opt}}}(s)}
\label{eq:rule_over_network}
\end{equation}
expressed as a percentage, indicating the additional relative scheduling costs incurred by using the advanced dispatching rule $\mu_{\text{rule,opt}}$ compared to our neural network $\mu_{\vartheta,\text{opt}}$ with trained parameters $\vartheta$. Our results show that our neural network significantly outperforms the heuristic method across all problem scenarios for up to 100 jobs and 10 machines, with the dispatching rule leading to 22.22\% higher average scheduling costs across the tested settings within this range. Performance tends to equalize around 12 machines. Interestingly, the largest advantages were not observed in the training scenario of 8 jobs and 4 machines but rather in ranges of 8 to 20 jobs and 4 to 6 machines as well as 15 to 30 jobs and 5 to 8 machines. Except for two configurations, in settings with up to 30 jobs and between 3 and 8 machines, the dispatching rule created schedules with average costs more than 20\% higher than those produced by our neural network, often exceeding 30\%.
\section{Limitations}

Despite being trained solely on scheduling problems limited to 8 jobs and 4 machines, our network has generalized remarkably well to much larger environments, handling configurations with over ten times the number of jobs it was originally trained on. With increased computational resources and by introducing GPUs to take advantage of the inherent parallelization facilitated by the networks architecture, the training set could be expanded to extend its generalization capabilities, particularly for accommodating higher numbers of machines.

Although we predefined parameter limitations in \cref{section:data}, our trained network can be applied directly to scheduling problems outside these ranges. Scaling all time-related values by a factor would neither alter cost relations between policies nor impact the network's state-action recommendations. Since we dynamically normalize inputs to between 0 and 1 on a state basis in \cref{section:data_preprocessing}, the model remains invariant to the absolute scale, enabling it to interpret time ratios outside those used during training. 

While our approach only addresses a specific, complex problem~(\ref{our_JSP}), it can readily be adapted to simpler environments by creating edge cases. For instance, machine or job tardiness can be disregarded by setting their respective weights to zero, and uniform-machines scheduling can be modeled by making processing times proportional. Additional constraints can be incorporated with very minimal changes to the network architecture. Future research could extend our framework by adding various constraints, making our specific problem an edge case of a more generalized solution, applicable to a vast range of problem settings. Retraining the model for different objective functions would be efficient due to the supervised learning setup, in contrast to standard rule-based algorithms, which often require customized solutions for each problem setting, or approaches relying on DRL, which is significantly slower to converge.

Though currently limited to deterministic settings, the methodology could be adapted for stochastic environments by incorporating expected processing times; for discrete distributions, separate vectors could represent potential processing times and their associated probabilities.

%Additionally, our approach is compatible with Deep Q-Learning, which could further improve performance or serve as a powerful initialization step for DRL models. As it operates on a state-by-state basis, the framework is also inherently suited for online scheduling.
\section{Conclusion}

Our developed network architecture demonstrated the necessary flexibility for unrelated-machines scheduling in a complex problem environment. It successfully handled the inherent variability not only in input and output sizes but also in input feature dimensions. Our network excelled in scheduling 8 jobs and 4 machines. Compared to the optimal policy, it induced less than 1\% additional costs per state decision and only 2.51\% across full schedules --- substantially lower than the 34.46\% incurred by the advanced dispatching method.
Despite being trained exclusively on these small scheduling environments, our network generalized remarkably well to significantly larger problems with up to 100 jobs and 10 machines, consistently outperforming the dispatching rule and achieving at least 10\% lower costs in most tested configurations. For all but two setups with 3 to 8 machines and up to 30 jobs, the dispatching rule produced schedules with at least 20\% and up to 38.27\% higher costs compared to our network. 

These results highlight our approach’s potential to provide the flexibility, adaptability, and generalizability essential for tackling deterministic scheduling problems on unrelated machines. Additional computational resources would enable efficient training on more diverse datasets, incorporating scheduling problems with higher job and machine counts to further improve the model’s generalization. Moreover, the framework's compatibility with DRL could be leveraged to further enhance the model's performance while starting from a strong basis. While this study focused solely on specific problem constraints, the adaptable nature of our architecture, combined with the rapid retraining enabled by our efficient supervised learning approach, invites future research to explore its efficacy under varied operational conditions and objective functions.
%While this study focused solely on specific problem constraints, the adaptable nature of our architecture invites future research to explore its efficacy under varied operational conditions and objective functions, for which the efficiency of our supervised learning approach enables rapid retraining.

\bibliographystyle{IEEEtran}
\bibliography{bib.bib}

\end{document}